# Resource-Performance Trade-offs in Intra-Aisle Robotic Picking: A Discrete-Event Simulation Comparison of Shuttle Configurations With and Without an Onboard Tote Lift

**Mohsen Reda Mahmoud Moursi***
Graz University of Technology, Graz, Austria
*Research conducted while the author was a Master's student.*
Email: mohsenredamahmoud@gmail.com
*Corresponding author: Mohsen Reda Mahmoud Moursi, mohsenredamahmoud@gmail.com

## Abstract

This study compares two implemented intra-aisle robotic-picking configurations through discrete-event simulation. Without an onboard tote lift, two level-captive shuttles serve the two tested levels; with the lift, one shuttle remains on its horizontal level and its onboard mechanism moves the order tote to allow picks from the adjacent upper level. Thirty configuration–task-size experiments (n = 1–15 locations; ten replications each) yield 300 replication-level cycle-time observations. Two-sided Welch tests are applied to the replication means, with Holm adjustment across the 15 comparisons. The onboard-tote-lift configuration has longer mean cycle time at all tested task sizes. At n = 15, the means are 299.477 and 333.231 s/cycle, respectively: a 33.753 s increase (11.27%). The corresponding cycle-time-derived item rates are 180.314 and 162.050 items/h (−10.13%). These rates are calculated as 3600n divided by the scenario-mean cycle time; they are not measurements of the combined concurrent capacity of two level-captive shuttles. Importantly, task locations are sampled from different eligible-level sets in the two configurations, so the result concerns the implemented configuration-and-task-generation packages, not the isolated effect of a five-second tote-lift operation under identical picked locations. The study quantifies a cycle-performance difference alongside a reduction from two dedicated shuttles to one within the tested two-level block; it does not quantify system-level maximum throughput, investment cost, or reliability.

## 1. Introduction

Automated warehousing increasingly combines storage, transport and picking functions to reduce manual travel and support short response times. The resulting systems are not interchangeable: performance depends on where item transfer occurs, which resources move totes or order containers, and whether horizontal and vertical movements are performed by dedicated or shared devices (de Koster et al., 2007; Boysen et al., 2019; Tappia et al., 2019). Shuttle-based storage and retrieval systems illustrate this design problem because the number of shuttles, their level coverage, lift architecture and control rules determine both resource requirements and achievable throughput (Eder, 2020; Ekren et al., 2024; Eder, 2024).

The system studied here is an intra-aisle robotic picking concept in which a mobile shuttle transports an order tote along the aisle and an integrated picking function transfers items at storage locations. This differs from conventional parts-to-picker architectures because the item transfer occurs inside the aisle. A related analytical formulation for a robotic order-picking shuttle was reported by Lerher (2018), and Lerher et al. (2025) later formulated expected cycle time and item throughput for a multi-command robotic picking cycle. The present article does not repeat the analytical-model correspondence assessment. Instead, it uses the simulation model to address a separate design decision: whether two adjacent storage levels should be served by separate level-captive shuttles or by one shuttle equipped with an onboard tote lift.

The two options represent a direct engineering trade-off. Without an onboard tote lift, each tested level has its own shuttle and track service. With an onboard tote lift, one shuttle remains on its assigned level while the tote lift moves the tote vertically relative to the shuttle so that picks can be made from both the shuttle level and the level above. The second configuration therefore reduces the dedicated shuttle count across the two tested levels from two to one, but adds lift operations and changes routing and controller behaviour. It must be assessed as a complete configuration change rather than as a five-second timing penalty in isolation.

Published shuttle-system studies show that architecture choices can trade investment, utilisation and throughput against one another. Tier-captive SBS/RS uses level-dedicated shuttles and separate lifts for tote movement, whereas tier-to-tier or movable-lift concepts redistribute transport resources across levels (Eder, 2020; Ekren et al., 2024). Eder (2024) further shows that lift capacity and shuttle-lift coordination are explicit performance variables. These studies provide the architectural context, but their transport processes and KPI boundaries are not identical to the intra-aisle item-picking process considered here. Their numerical results are therefore not used as matched benchmarks.

The research question is: Under the defined two-level benchmark and its configuration-dependent task generation, how do replication-mean picking-cycle times and cycle-time-derived item rates differ between two level-captive shuttles and one shuttle with an onboard tote lift? The study also records the change in dedicated shuttle count. It does not estimate the aggregate concurrent capacity of the two-shuttle arrangement. The contribution of the article is a bounded configuration comparison with an explicit KPI boundary and independently reproducible replication-level statistical results. The comparison was originally developed as Study B of the author's Master's thesis (Moursi, 2026) and is presented here as a focused article rather than a claim of previously unreported simulation experiments.

## 2. Related Work and Study Positioning

### 2.1 Automated storage and picking architectures

Warehouse architecture determines which movements and resource interactions enter a performance model. In tier-captive SBS/RS, shuttles are dedicated to storage tiers and vertical tote movement is provided by lifts. Lerher et al. (2015) analyse shuttle travel times in a conventional storage/retrieval setting, while Eder (2020) develops a performance model for tier-captive systems with class-based storage. Tappia et al. (2019) consider integrated storage and picking, emphasizing the influence of buffers and resource interactions. These studies motivate explicit modelling of motion and handling but do not provide directly matched numerical benchmarks for the present in-aisle item-picking cycle. A systematic review by Saini et al. (2025) documents the wider use of simulation to evaluate warehouse processes and picking operations; it does not establish a matched result for this particular shuttle comparison.

Resource consolidation and shuttle sizing have been studied in other architectures. Küçükyaşar et al. (2021) compare tier-captive and tier-to-tier SBS/RS cost and performance, and Ha and Chae (2019) develop a decision model for shuttle numbers in tier-to-tier SBS/RS. Ekren et al. (2024) compare tier-captive SBS/RS with an AVS/RS using movable lifts; Eder (2024) analyses systems with multiple-capacity lifts. Recent studies additionally examine tier-to-tier operation and sequencing (Xu et al., 2025; Chen and Yang, 2026), as well as simulation-based time and energy implications (Ferrari et al., 2025). In those tier-to-tier configurations, the shuttle can move between tiers using separate infrastructure. Here the shuttle remains on its assigned horizontal level and only the carried tote is raised for adjacent-level picking. The different physical processes and output measures prevent treating results from those publications as direct numeric comparators.

### 2.2 Intra-aisle robotic picking

The investigated concept differs from standard shuttle storage/retrieval transport because the picking function travels to each selected storage location. Lerher (2018) proposed an analytical description of a robotic order-picking shuttle carrier; Lerher et al. (2025) considered a multi-command picking cycle with varying numbers of locations. The present comparison retains the intra-aisle picking process but changes the complete shuttle allocation, eligible picking levels and associated routing/controller behaviour between the two configurations. It is not a lift-service-time-only experiment.

The present comparison is deliberately narrow. It does not claim that an onboard tote lift is universally preferable or inferior to level-captive shuttles. Instead, it quantifies what is gained and lost when one shuttle is allowed to serve two adjacent storage levels in the implemented intra-aisle robotic-picking process. The result is intended as design evidence for the tested model boundary, not as a direct cost-benefit result for a commercial warehouse.

## 3. System Definition and Compared Configurations

### 3.1 Common process boundary

Both configurations use the same underlying intra-aisle picking process. A shuttle transports an order tote horizontally along the aisle, visits selected storage locations, executes tote handling and item picking, and returns to its home position after the required picks. The configured item-picking time is 10.0 s per picked item. Tote handling is 3.4 s per applicable pickup action and 3.4 s per applicable set-down action. The tested shuttle speed is 2.5 m/s and acceleration 1.5 m/s². Replenishment is disabled in the paper-benchmark comparison; this does not imply that the wider Plant Simulation model lacks replenishment processes.

The reported clean cycle-time boundary includes horizontal travel, tote handling, item picking and, when used in the onboard-tote-lift configuration, tote-lift service. The configured shuttle tote-lift input is 5.0 s for an applied level-transition operation. The shuttle does not move vertically. The wider model has common entry/exit interfaces, but the present outcome is a picking-cycle measure, not a directly counted number of completed orders at the overall exit.

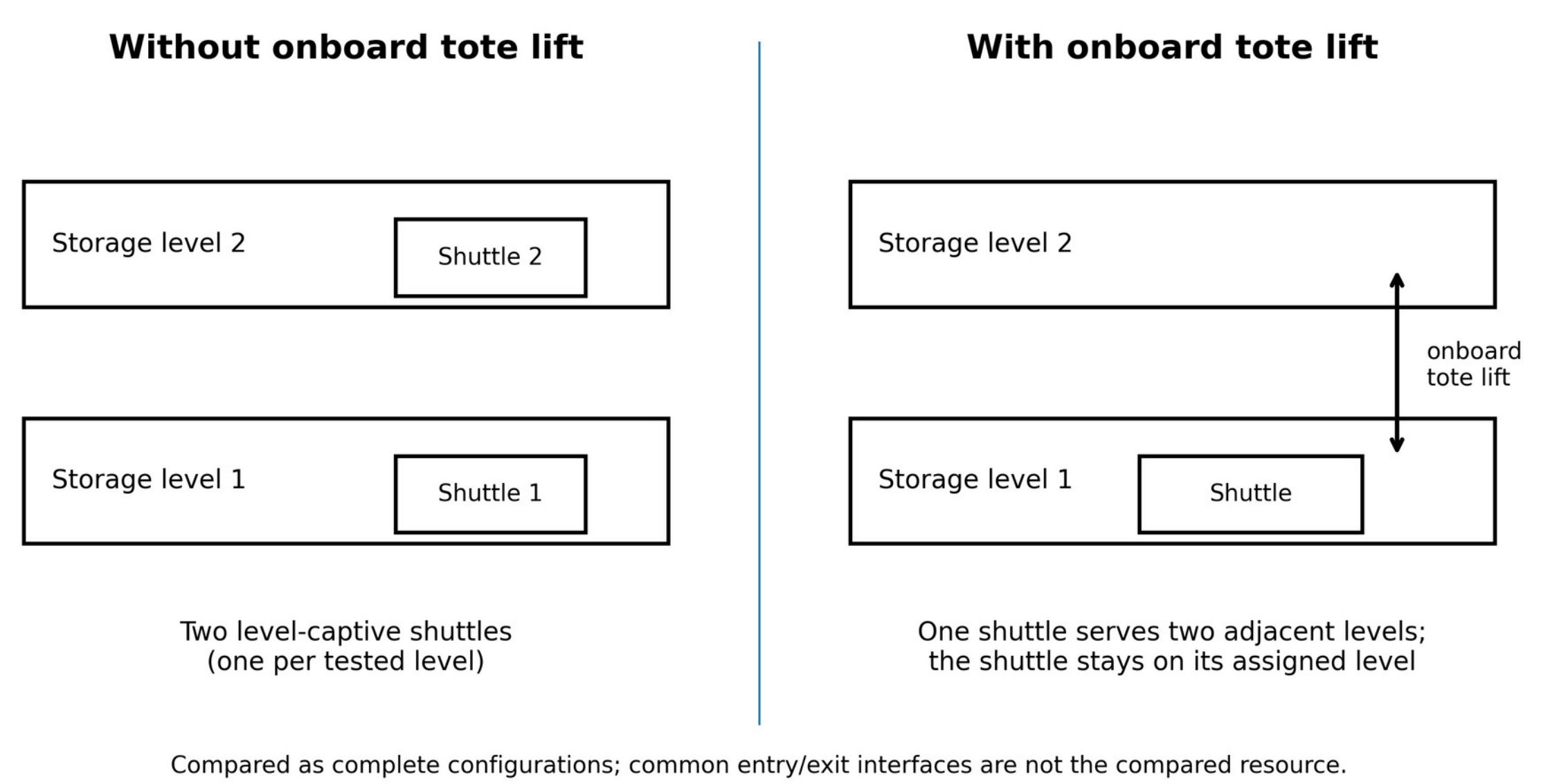

Figure 1. Compared shuttle configurations. The onboard tote lift changes level coverage while the shuttle itself remains on its assigned level.

### 3.2 Configuration A: without onboard tote lift

In the configuration without an onboard tote lift, the two tested storage levels are served by separate level-captive shuttles. Each shuttle remains on its assigned level and serves tasks from that level only. In implementation terms, Var_CoveredRows = 1 and Var_PaperMultiLevel = 0 define this arrangement. Across the two tested levels, the configuration therefore uses two dedicated shuttles.

### 3.3 Configuration B: with onboard tote lift

In the onboard-tote-lift configuration, one shuttle serves two adjacent storage levels. The shuttle itself remains on its assigned horizontal level; the tote lift moves the tote relative to the shuttle when access to the adjacent upper level is required. Var_CoveredRows = 2 and Var_PaperMultiLevel = 1 enable the corresponding task-generation and control logic. Level transitions invoke Var_ShuttleLiftTime, while consecutive picks on the same level do not require an additional tote-lift transition.

**Table 1. Structural difference between the two compared configurations.**

| Dimension | Without onboard tote lift | With onboard tote lift | Held/common interpretation |
|---|---|---|---|
| Dedicated shuttles across two tested levels | 2 | 1 | Resource-count trade-off |
| Level coverage per shuttle | One level | Two adjacent levels | Same intra-aisle picking function |
| Shuttle vertical movement | None | None | Shuttle stays on assigned horizontal level |
| Tote vertical movement on shuttle | Not required | Onboard tote lift when upper-level access is needed | 5.0 s service-time input when applied |
| Tote handling / item picking | 3.4 s pick-up/set-down; 10.0 s item pick | Same | Held common |
| Replenishment in controlled comparison | Disabled | Disabled | Not part of Study B cycle comparison |

## 4. Experimental Design and Analysis

### 4.1 Discrete-event simulation experiment

The retained model was inspected using Siemens Tecnomatix Plant Simulation 24.4.9.3160; the precise software version used during the historical Study B runs was not independently established. Study B includes 15 numbers of visited picking locations (n = 1–15) crossed with the two configurations, producing 30 experiments and ten simulation replications per experiment (300 replication records). Configuration A has Var_CoveredRows = 1 and Var_PaperMultiLevel = 0, while configuration B has Var_CoveredRows = 2 and Var_PaperMultiLevel = 1. Benchmark mode is enabled (Var_PaperBenchmark = 1). The two configurations share the geometric, movement and handling inputs listed in Table 2a. The order cap Var_MaxOrders = 500 is a configured input, not proof of 500 completed or accepted picking cycles. Replication numbers were not treated as statistically paired.

For the reported benchmark task generator, storage locations with positive stock are eligible within the designated aisle. Each order's n selected locations are sampled without replacement. Without the onboard lift, eligible locations are on the assigned shuttle level; with the lift, eligibility includes the assigned level and the adjacent upper level (thesis Appendix A.1). Thus equal n does not establish identical physical picking locations, level mixes or travel demands. The preserved cleaned export has all ExpID–RepNo combinations, no duplicate pairs, and complete cycle-time fields. It records 8,531 accepted complete cycles over the 300 replication summaries, zero invalid cycles and 242 end-cutoff exclusions. Acceptance counts vary from 10 to 110 per replication. Of the 300 retained replication rows, 284 have status OK and 16 have status cycle_time_out_of_tolerance. The available EndSim program text defines this label through comparison with an analytical cycle-time reference, separately from cycle-validity rules. All 300 exported labels were reproduced from the retained values and that rule, but the executable revision used in the historical runs has not been independently verified.

Under the available EndSim program text, a picking cycle is accepted when it contains exactly n Pick rows, a picked quantity of n, one DropOff row and a positive recorded cycle time; paper-benchmark mode additionally requires n distinct stock-location keys. A cycle with Pick rows but no DropOff at the run cutoff is recorded as excludedEndCutoff, not as an invalid completed cycle. Separately, EndSim labels a replication cycle_time_out_of_tolerance when the absolute percentage difference between the cleaned simulated cycle time and the analytical reference exceeds 5%. All 16 such labels have been reproduced from the retained replication means: WITHOUT n = 1 (10 records), WITH n = 1 (four), and WITHOUT n = 2 (two). The 16 records are retained because the flag identifies analytical-reference disagreement, not a failure of the completed-cycle acceptance checks. The single-location case is separately addressed in the source thesis (§5.5.3), which reports a simulated cycle time of 33.405 s against the analytical 37.333 s for the corresponding 30 m reference. These data do not by themselves isolate the physical cause of the discrepancy. The experimental description follows the STRESS reporting principles (Monks et al., 2019).

**Table 2. Experimental design and statistical analysis.**

| Item | Study design |
|---|---|
| Experimental factor | Shuttle configuration: without vs with onboard tote lift |
| Task size | n = 1-15 visited picking locations |
| Experiments | 30 (15 task sizes x 2 configurations) |
| Replications | 10 per experiment; 300 complete replication-level observations |
| Primary KPI | Replication mean clean cycle time, s/cycle |
| Derived KPI | Item throughput = 3600n / scenario-mean cycle time, items/h |
| Inferential comparison | Two-sided Welch independent-samples test at each n |
| Multiplicity control | Holm adjustment across 15 task-size comparisons |
| Interval estimate | 95% Welch confidence interval for with-minus-without cycle-time difference |

Table 2a. Fixed Study B benchmark inputs and interpretation (thesis Appendix A.5).

| Setting | Value and interpretation |
|---|---|
| Geometry: tested rack length | 30 m; 60 columns; two adjacent tested storage levels |
| Shuttle motion | v_max = 2.5 m/s; acceleration = 1.5 m/s² |
| Picking / handling | Item pick = 10.0 s; tote pickup = 3.4 s; tote setdown = 3.4 s |
| Onboard tote lift | Var_ShuttleLiftTime = 5.0 s for applicable service |
| Experiment factor | n = 1–15 visited picking locations; two configurations |
| Level coverage without lift | Var_CoveredRows = 1; Var_PaperMultiLevel = 0 |
| Level coverage with lift | Var_CoveredRows = 2; Var_PaperMultiLevel = 1 |
| Task / stock setup | Var_PaperBenchmark = 1; positive-stock candidates; no duplicate location in one order; benchmark occupancy 100%; replenishment disabled |
| Configured order cap | Var_MaxOrders = 500 (not a confirmed accepted-cycle count) |
| Statistical repetitions | 10 per configuration × n cell; 300 total |

## 4.2 Performance measures

For each replication, clean cycle time is the mean of its accepted, completed picking cycles. The scenario result CT-04 is the unweighted mean of the ten replication means CT-03; individual cycles are not regarded as independent inferential units. At each n, the principal difference is ΔCT = CTwith − CTwithout, so a positive difference means a longer cycle with the onboard tote lift. The supplied cleaned file retains accepted-cycle counts and excluded-end-cutoff counts at replication level. The current ExperimentManager source was additionally inspected: its inherited DefSeed4Run method assigns EventController.RandomNumbersVariant = ObservNo when UserSeedValueMethod is blank, and the current override value is blank. Thus the current default rule uses the replication index (1–10) as the random-number variant. However, the inspected saved model currently holds Study C settings (eight experiments, Var_PaperBenchmark = 0), not the historical Study B campaign. The original Study B stopping time and whether exactly this source and seed rule generated the archived 300 records cannot be recovered from the cleaned CSV alone; no historical settings are asserted without supporting records.

The secondary, derived item rate is TH-03(n) = 3600n / CT-04(n) items/h, where n is the number of picked items in the measured multi-command cycle and CT-04 is the unrounded scenario-mean cycle time in seconds. TH-03 is a mathematical transformation of the cycle-time result; it does not measure aggregate system capacity under simultaneous demand served by the two level-captive shuttles. Both shuttles exist in configuration A, but the present TH-03 data cannot establish their combined concurrent production capacity. No 2× multiplier is applied to TH-03 because shared resources and operational overlap would require a different, observed system-level metric.

At each task size, the ten replication means for the two configurations are compared by a two-sided independent-samples Welch t-test, without an equal-variance assumption and without assuming common random numbers. The Holm method adjusts the set of 15 p-values to control familywise error. The displayed 95% Welch confidence intervals are pointwise, not simultaneously multiplicity-adjusted intervals. Replication-level output analysis is consistent with established guidance on statistical analysis of simulation results (Law, 2020). The 15 numerical comparisons and adjusted p-values can be recomputed from the author-retained replication CSV and analysis code; those files are not included in this PDF-only submission.

## 5. Results

### 5.1 Cycle-time comparison

The onboard-tote-lift configuration has the higher mean cycle time at every tested task size. Mean cycle time rises from 33.405 to 38.293 s/cycle at n = 1, from 171.123 to 193.544 s/cycle at n = 8, and from 299.477 to 333.231 s/cycle at n = 15. Figure 2 shows that the gap persists over the complete n = 1-15 range.

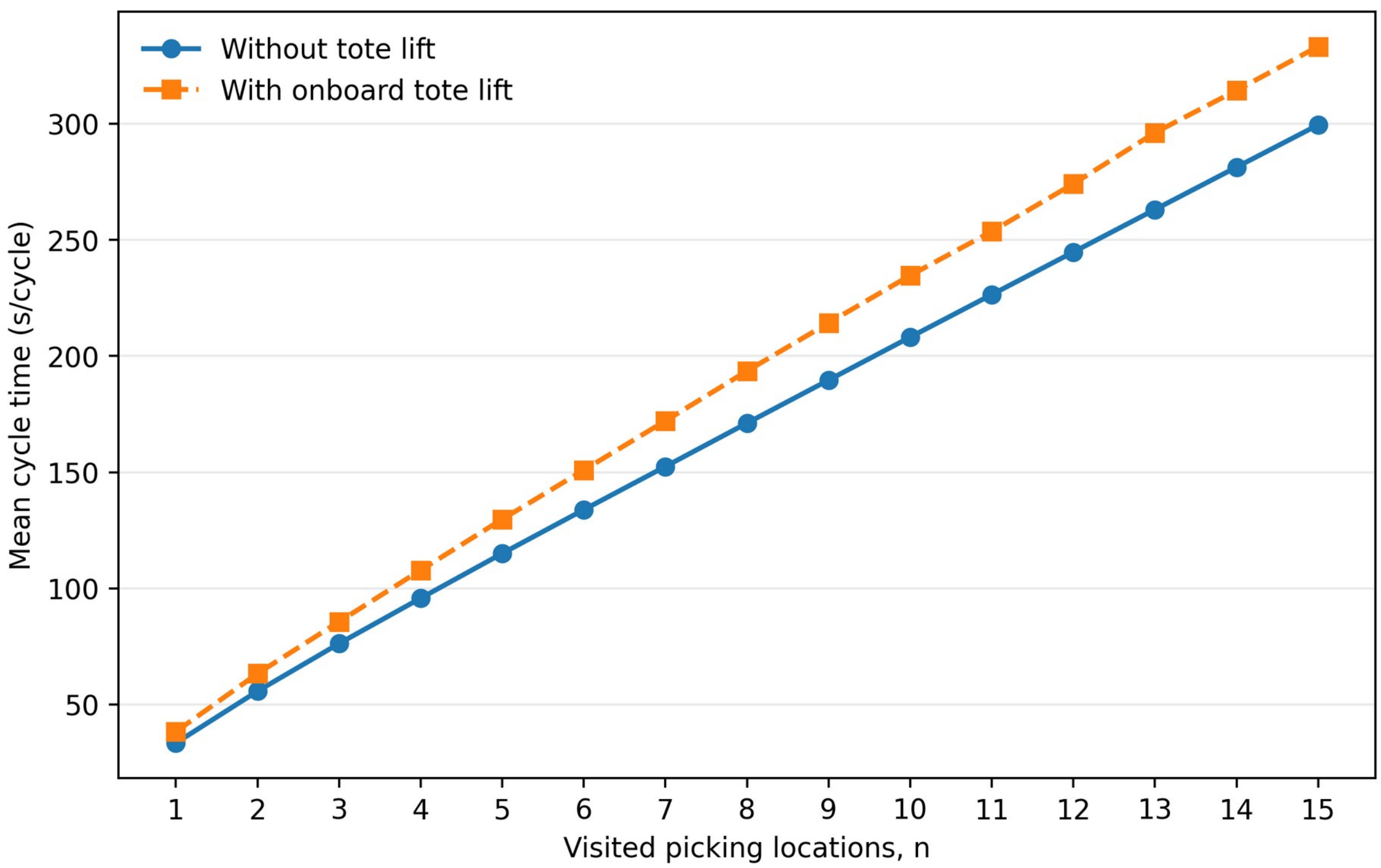

Figure 2. Mean cycle time by shuttle configuration and number of visited picking locations.

The absolute with-minus-without difference generally grows with task size but is not strictly monotone: it is 4.888 s at n = 1 (pointwise 95% Welch CI 4.168–5.609 s), 22.421 s at n = 8 (20.924–23.918 s), and 33.753 s at n = 15 (31.540–35.967 s). The difference at n = 14 is slightly below that at n = 13. All 15 pointwise Welch intervals exclude zero, and all 15 tests remain significant following Holm adjustment of the p-values (each adjusted $p < 0.001$). Figure 3 shows the estimated differences with pointwise intervals; the intervals themselves are not familywise-adjusted.

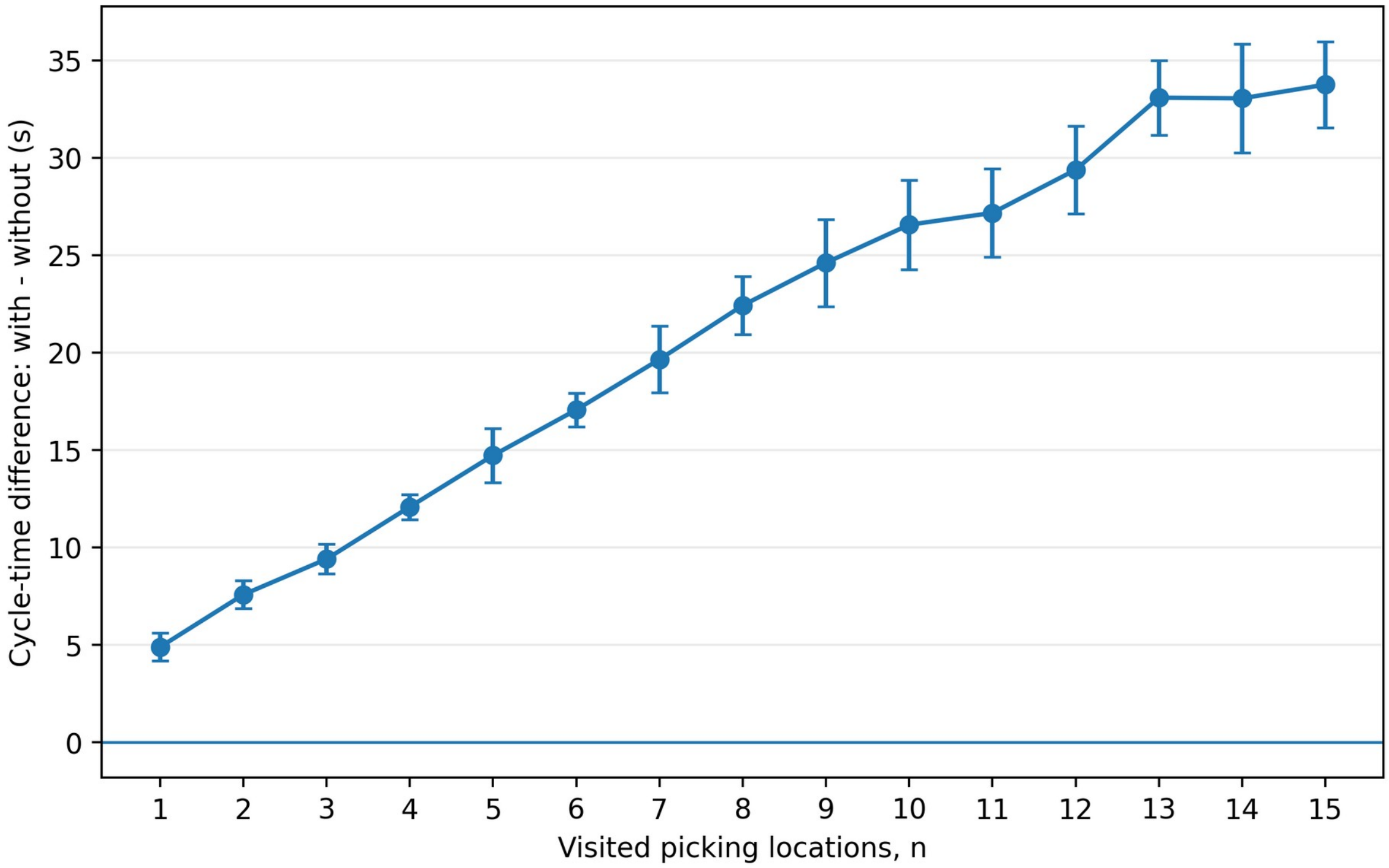

Figure 3. Cycle-time difference (with minus without onboard tote lift) and pointwise 95% Welch confidence intervals; Holm adjustment applies to p-values, not to the displayed intervals.

Expressed relative to the without-tote-lift configuration, the cycle-time increase ranges from 11.27% to 14.63% across the 15 task sizes, with an unweighted mean of 12.67%. The percentage penalty becomes somewhat smaller at the highest task sizes even though the absolute difference in seconds increases.

### 5.2 Derived item throughput

Because both configurations pick the same number n of items per measured cycle, the longer cycle time in the onboard-tote-lift configuration produces lower derived item throughput at every tested task size. At $n = 1$, derived throughput is 107.769 items/h without the tote lift and 94.012 items/h with it. At $n = 8$, the corresponding values are 168.300 and 148.803 items/h. At $n = 15$, they are 180.314 and 162.050 items/h.

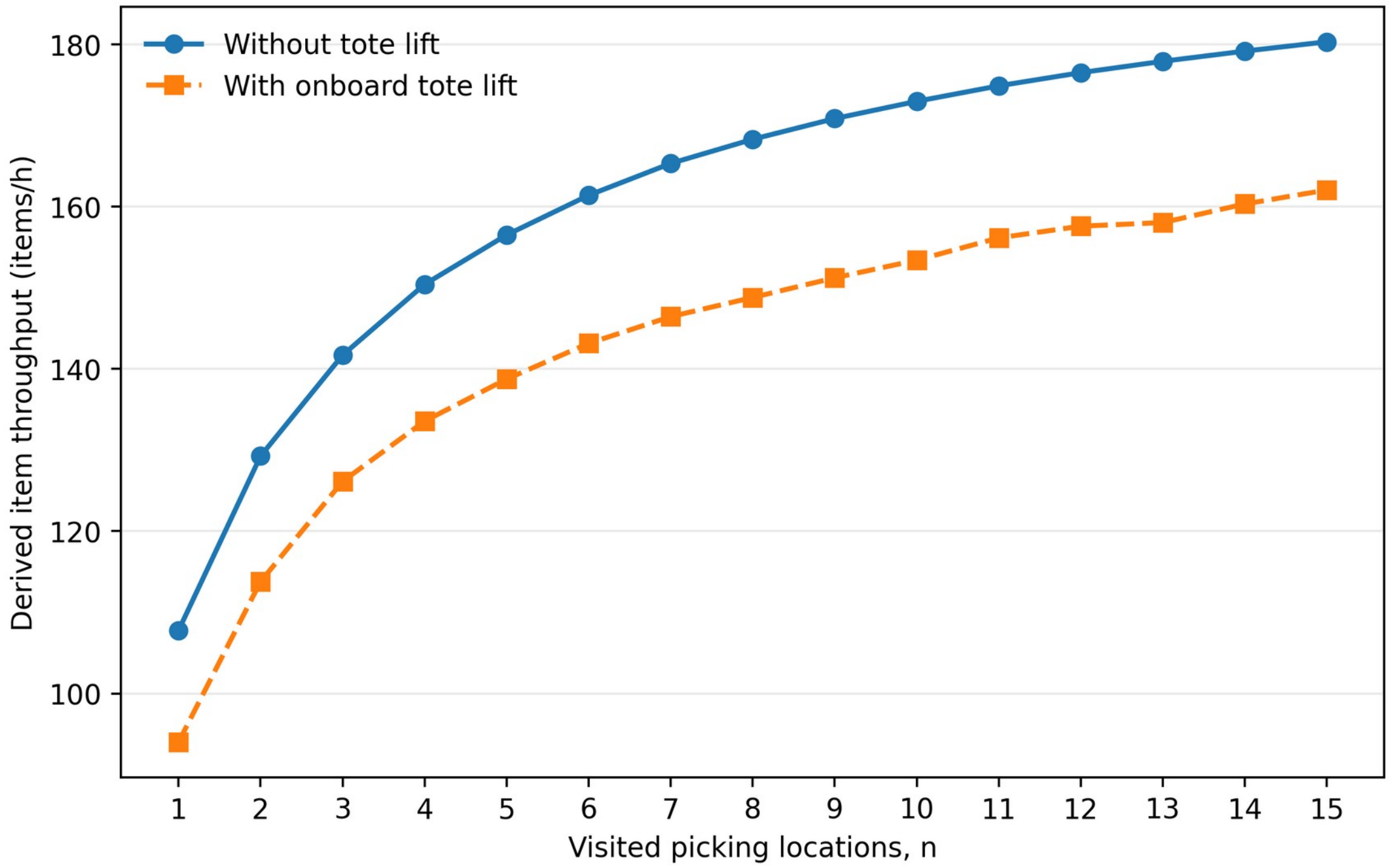

Figure 4. Cycle-time-derived item rate (items/h) for each configuration: 3600n / scenario-mean cycle time. These values are not measured two-shuttle concurrent system capacity.

Across n = 1–15, the relative change in the cycle-time-derived rate ranges from −12.77% to −10.13% and averages −11.24% over the 15 task sizes. These changes follow algebraically from the measured cycle-time differences at fixed n and are not a separate experimental estimate of simultaneous system throughput. They must not be interpreted as the percentage change in combined two-level order-completion capacity.

### 5.3 Consolidated configuration comparison

**Table 3. Cycle time, derived throughput and configuration effect across n = 1-15.**

| n | CT without (s) | CT with (s) | ΔCT (s) | 95% CI ΔCT | CT change | TH without | TH with | TH change |
|---|---|---|---|---|---|---|---|---|
| 1 | 33.405 | 38.293 | +4.888 | 4.17-5.61 | +14.63% | 107.769 | 94.012 | −12.77% |
| 2 | 55.703 | 63.279 | +7.576 | 6.86-8.29 | +13.60% | 129.256 | 113.782 | -11.97% |
| 3 | 76.219 | 85.623 | +9.404 | 8.64-10.17 | +12.34% | 141.696 | 126.134 | -10.98% |
| 4 | 95.733 | 107.816 | +12.083 | 11.44-12.73 | +12.62% | 150.419 | 133.561 | -11.21% |
| 5 | 115.008 | 129.730 | +14.722 | 13.33-16.12 | +12.80% | 156.511 | 138.750 | -11.35% |
| 6 | 133.832 | 150.895 | +17.062 | 16.19-17.93 | +12.75% | 161.396 | 143.146 | -11.31% |
| 7 | 152.429 | 172.080 | +19.651 | 17.94-21.36 | +12.89% | 165.323 | 146.444 | -11.42% |
| 8 | 171.123 | 193.544 | +22.421 | 20.92-23.92 | +13.10% | 168.300 | 148.803 | -11.58% |
| 9 | 189.638 | 214.249 | +24.611 | 22.37-26.85 | +12.98% | 170.852 | 151.226 | -11.49% |
| 10 | 208.118 | 234.678 | +26.560 | 24.27-28.85 | +12.76% | 172.979 | 153.402 | -11.32% |
| 11 | 226.431 | 253.601 | +27.171 | 24.91-29.43 | +12.00% | 174.888 | 156.151 | -10.71% |
| 12 | 244.743 | 274.125 | +29.382 | 27.13-31.63 | +12.01% | 176.512 | 157.592 | -10.72% |
| 13 | 263.052 | 296.135 | +33.083 | 31.17-35.00 | +12.58% | 177.912 | 158.036 | -11.17% |
| 14 | 281.305 | 314.353 | +33.048 | 30.26-35.84 | +11.75% | 179.165 | 160.329 | -10.51% |
| 15 | 299.477 | 333.231 | +33.753 | 31.54-35.97 | +11.27% | 180.314 | 162.050 | -10.13% |

All 15 Welch comparisons have Holm-adjusted $p < 0.001$. Table 3 reports cycle-time-derived rates in items/h, not measured concurrent system capacity. The confidence intervals shown in Table 3 are pointwise 95% Welch intervals. Exact test statistics, degrees of freedom and Holm-adjusted p-values are retained in the author's statistical analysis files; they are not part of this PDF-only preprint.

## 6. Discussion

### 6.1 Resource-performance trade-off

The central observed finding is a change in cycle-level performance alongside a reduction in dedicated shuttle resources, not a directly measured maximum-capacity trade-off. The onboard-tote-lift arrangement uses one shuttle across the two tested levels instead of two level-captive shuttles: a 50% reduction in shuttle count for this service block. Its measured mean picking-cycle times are 11.27–14.63% longer, with mathematically derived per-cycle item rates 10.13–12.77% lower. Because the two-shuttle baseline can operate concurrently, neither these percentages nor the displayed rates measure the combined available capacity of both shuttles. Whether one shared shuttle is sufficient for a specified overall demand requires a system-level completed-output study and economic inputs beyond the present Study B results.

The difference must not be attributed entirely to the nominal five-second tote-lift input. Changing configuration also changes eligible storage levels, the set of sampled locations, route opportunities and controller behaviour. Tote-lift events arise only when the controller requires a level change, so actual lift counts depend on the realised task sequence. The available 300-row replication summary contains mean cycle time but not per-cycle lift counts or component-time decomposition. No numerical apportionment to lift service, horizontal travel, or task mix is therefore supported by these results.

This interpretation is consistent with the wider shuttle-system literature, in which performance depends on how horizontal and vertical movements are distributed among shuttles and lifts. Tier-captive systems dedicate more horizontal resources to individual levels, while tier-to-tier or movable-lift approaches share transport resources more broadly and can reduce investment or resource count at the cost of additional coordination and travel (Eder, 2020; Ekren et al., 2024). The present system differs because the shuttle itself remains on its assigned level and the onboard mechanism moves the tote for adjacent-level picking, but the underlying allocation trade-off is similar: resource sharing changes service time and control requirements.

### 6.2 Effect of task size

The observed absolute cycle-time difference generally increases with n, rising from 4.888 s at n = 1 to 33.753 s at n = 15, while remaining slightly non-monotone around n = 13–14. More visited locations provide more opportunities for level changes and longer travel, but realised lift-operation counts and route-length differences were not available in the replication summary. Therefore, these candidate mechanisms cannot be quantitatively separated or assigned causal shares. The relative difference remains within approximately 11–15% across the studied values of n; this describes the simulated outcomes rather than a universal scaling law.

The derived throughput curves approach a flatter region as n grows because more items are completed per cycle while the fixed start/return components are distributed over more picks. Both configurations benefit from this batching effect. The onboard-tote-lift configuration remains below the level-captive configuration throughout the tested range, but its relative throughput penalty falls to about 10.13% at n = 15.

### 6.3 Design interpretation and use

For early-stage design, these results provide bounded evidence about per-cycle speed at the fixed Study B settings and about the number of shuttles allocated to the tested two-level block. They do not establish whether either configuration meets a target demand expressed as aggregate completed items or orders per hour. That decision requires a separate completed-output rate measured over a specified simulation horizon under comparable concurrent demand, plus assessment of common resources. Equipment pricing, availability, reliability, maintenance and energy must also be considered before inferring any economic advantage from the lower shuttle count.

The configuration choice also has resilience implications that are outside the measured KPIs. Two level-captive shuttles provide resource redundancy across the two levels, whereas a single shared shuttle represents a more concentrated failure point. Conversely, the one-shuttle configuration uses fewer moving resources. Neither effect is quantified here because failures and maintenance were not included in the model. They should therefore be evaluated explicitly before the resource-count reduction is treated as a life-cycle advantage.

## 7. Limitations

The results apply only to the tested two-level, 30 m benchmark with one travel-speed/acceleration setting, a five-second nominal tote-lift service input and n = 1–15 visited locations. They are simulation outcomes without calibration to an operating warehouse. The configurations are not matched to identical sampled storage locations because eligibility differs by level coverage; the design therefore estimates an implemented configuration-and-task-generation contrast, not the isolated effect of enabling a lift under fixed requests. The secondary rate is derived from cycle time and does not measure concurrent two-shuttle system capacity. The dataset also does not provide realised lift counts or a component-time breakdown. Failures, maintenance, costs, energy, larger level sets and different workload mixes were not evaluated, so system-capacity and economic generalisations are not supported.

The 300-row cleaned replication export is complete with respect to ExpID × RepNo and includes 242 end-cutoff exclusions and 16 cycle_time_out_of_tolerance labels, with zero invalidCycles. The available EndSim program text specifies a tolerance flag when the absolute discrepancy from the analytical cycle-time reference exceeds 5%; reconstruction from the cleaned values matches all 300 exported labels, but does not independently validate the historical run execution. WITHOUT at n = 1 is systematically flagged in ten of ten replications; the source thesis independently discusses n = 1 as a special case in its analytical-versus-simulation comparison. These labels identify reference-model disagreement, not automatically invalid completed cycles. All 300 replication means are retained in the primary Welch/Holm analysis. Inspection of the current ExperimentManager shows a blank custom-seed override and an inherited default rule mapping the random-number variant to the replication index. Because that saved configuration currently represents Study C, it is evidence for the present setup rather than definitive proof of Study B's historical seed and stop configuration. The cleaned export is also distinct from a complete raw event-log archive, and the precise executable-source-to-export provenance remains incompletely documented. Separately retained statistical support files reproduce the statistical results from the clean records without inventing missing execution settings.

## 8. Conclusion

This article presents a focused comparison of two simulated intra-aisle picking configurations for n = 1–15 visited picking locations. The baseline employs two level-captive shuttles, one on each tested level. The alternative employs one shuttle that remains on its horizontal level while its onboard tote lift provides picking access to the adjacent upper level. In the implemented benchmark, the resource count across the tested two-level block changes from two shuttles to one; the underlying eligible sampling population and controller logic change as part of the configuration.

Across ten replications per configuration and task size, the onboard-lift configuration has a longer scenario-mean picking-cycle time at every tested n. All 15 two-sided Welch tests remain significant after Holm p-value correction; the displayed 95% Welch intervals are pointwise. At n = 15, the cycle-time means are 299.477 and 333.231 s, a difference of 33.753 s (11.27%). The corresponding 180.314 and 162.050 items/h values are derived from cycle time and do not represent combined system capacity from concurrently operating shuttles.

The defensible design implication is limited to the examined benchmark: reducing dedicated shuttle count is associated with longer measured picking cycles in the implemented configuration packages. The experiment does not establish the aggregate simultaneous output of the two-shuttle architecture, the separate causal contribution of the tote lift, or a cost-optimal design. Future study could evaluate matched location-level demand, end-to-end concurrently completed output and component-time decomposition across varied lift and motion settings. The present contribution is the transparent, replication-based comparison under its stated cycle-time boundary, with the reported summary statistics backed by author-retained replication records and reproduction code.

## Data availability

The statistical analysis uses a cleaned 300-row replication-level dataset, a 30-scenario summary, and 15 Welch/Holm comparisons. The author retains the cleaned replication records, scenario summary, comparison outputs and analysis code used to reproduce the reported contrasts. These files are not provided with this PDF-only preprint, and no public data repository is linked. The proprietary Plant Simulation model is likewise not included. The original historical run-control settings, executable-source-to-export linkage and complete raw event logs have not been independently verified.

## Origin of the study

The Study B configuration comparison and its numerical results originate from the author's 2026 Master's thesis, Modelling and Simulation of Intra-Aisle Picking Operations in Automated Warehousing (Moursi, 2026, Section 5.7). This paper presents a focused, bounded interpretation of that study; it does not portray the simulations as newly conducted experiments. Moursi and Stohy (2026, arXiv:2610.25967) address analytical-model correspondence rather than the configuration contrast reported here. Figures 1–4 are adapted from the author's prior research materials.

## Acknowledgments

I would like to thank Ahmed Stohy for his advice and support in preparing this paper for publication.

## Declaration of generative AI use

The simulation model and study data originate from the author's Master's research. OpenAI ChatGPT (GPT-6) was used to assist with substantial manuscript drafting and revision, statistical-code review, and preparation of publication materials. The author is responsible for verifying the final scientific statements, calculations, references, and figures.